\pdfoutput=1
\documentclass[11pt]{article}

\usepackage[]{ACL2023}

\usepackage{times}
\usepackage{latexsym}
\usepackage{pifont}% http://ctan.org/pkg/pifont

\usepackage[T1]{fontenc}
\usepackage{latexsym}
\usepackage{amsmath}
\usepackage{tabularx}
\usepackage[font=small,skip=2pt]{caption}
\usepackage{placeins}
\usepackage{multirow}
\usepackage{amssymb}
\usepackage{graphicx}
\usepackage{fixltx2e}
\usepackage{tabularx}
\usepackage{color, colortbl}
\usepackage{booktabs, siunitx}
\usepackage{etoolbox}
\newcolumntype{d}[1]{D{.}{.}{#1}}
\newcolumntype{s}[1]{S{.}{.}{#1}}

\newcommand\mc[1]{\multicolumn{1}{c}{#1}} % handy shortcut macro

\usepackage{siunitx}
	
\definecolor{LightCyan}{rgb}{0.88,1,1}
\definecolor{lc3} {gray}{.65}
\definecolor{lc2} {gray}{.75}
\definecolor{lc1} {gray}{.85}
\definecolor{lc0} {gray}{.95}
\newcommand*{\SuperScriptSameStyle}[1]{%
  \ensuremath{%
    \mathchoice
      {{}^{\displaystyle #1}}%
      {{}^{\textstyle #1}}%
      {{}^{\scriptstyle #1}}%
      {{}^{\scriptscriptstyle #1}}%
  }%
}

\newcommand*{\oneS}{\SuperScriptSameStyle{*}}
\newcommand*{\twoS}{\SuperScriptSameStyle{*}}
\newcommand*{\threeS}{\SuperScriptSameStyle{*}}

\newcommand{\Sref}[1]{\S\ref{#1}}

\newcolumntype{d}[1]{D{.}{.}{#1}}

\usepackage[utf8]{inputenc}

\usepackage{microtype}

\title{Annotating Mentions Alone Enables Efficient Domain Adaptation for Coreference Resolution}
\author{Nupoor Gandhi, Anjalie Field, Emma Strubell \\
  Carnegie Mellon University  \\
  \texttt{\{nmgandhi, anjalief, estrubel\}@cs.cmu.edu} }

\begin{document}
\maketitle

\begin{abstract}
Although recent neural models for coreference resolution have led to substantial improvements on benchmark datasets, transferring these models to new target domains containing out-of-vocabulary spans and requiring differing annotation schemes remains challenging. Typical approaches involve continued training on annotated target-domain data, but obtaining annotations is costly and time-consuming. 
We show that annotating mentions alone is nearly twice as fast as annotating full coreference chains. 
Accordingly, we propose a method for efficiently adapting coreference models, which includes a high-precision mention detection objective and requires annotating only mentions in the target domain. Extensive evaluation across three English coreference datasets: CoNLL-2012 (news/conversation), i2b2/VA (medical notes), and previously unstudied child welfare notes, reveals that our approach facilitates annotation-efficient transfer and results in a 7-14\% improvement in average F1 without increasing annotator time\footnote{Code is available at \url{https://github.com/nupoorgandhi/data-eff-coref}}.
% In this work, we show that adapting mention detection is the key component to successful domain adaptation of coreference models, rather than antecedent linking. 
% We also show annotating mentions alone is nearly twice as fast as annotating full coreference chains. 
% Based on these insights, we propose a method for efficiently adapting coreference models, which includes a high-precision mention detection objective and requires annotating only mentions in the target domain. 
\end{abstract}

\section{Introduction}
\begin{figure}[ht]
% \vspace{1ex}%
    \centering
    \includegraphics[width=.47\textwidth]{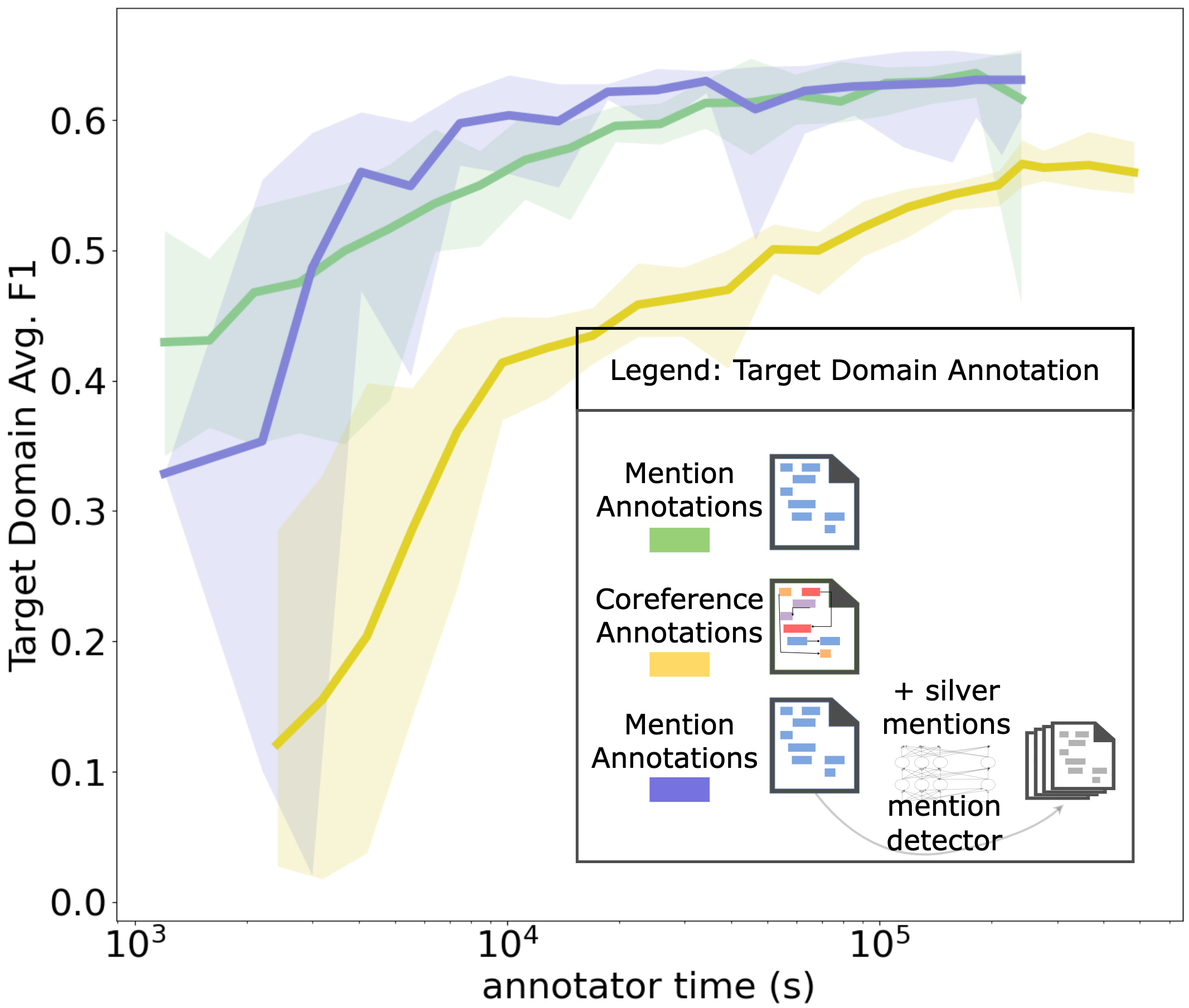}
    \caption{Model coreference performance (avg F1) as a function of continued training on limited target domain data requiring varying amounts of annotator time. The source domain is news/conversation (OntoNotes) and the target domain is medical notes (i2b2/VA). Using our method to adapt coreference models using only mentions in the target domain, we achieve strong coreference performance with less annotator time. }
    \label{fig:method-summary}
\end{figure}

Neural coreference models have made substantial strides in performance on standard benchmark datasets such as the CoNLL-2012 shared task, where average F1 has improved by 20\% since 2016 \citep{durrett-klein-2013-easy,dobrovolskii-2021-word, kirstain-etal-2021-coreference}. Modern coreference architectures typically consist of an encoder, mention detector, and antecedent linker. All of these components are optimized \textit{end-to-end}, using only an antecedent linking objective, so expensive coreference chain annotations are necessary for training \citep{aralikatte-sogaard-2020-model, li-etal-2020-neural}.

% Neural coreference models have made substantial strides in performance on standard benchmark datasets such as the CoNLL-2012 shared task, where average F1 has improved by 20\% since 2016 \citep{durrett-klein-2013-easy, xu-choi-2020-revealing,dobrovolskii-2021-word}. Modern coreference architectures typically consist of an encoder, mention detector, and antecedent linker. All of these components are optimized \textit{end-to-end}, using only an antecedent linking objective \citep{kirstain-etal-2021-coreference,dobrovolskii-2021-word}, so expensive coreference chain annotations are necessary for training \citep{aralikatte-sogaard-2020-model, li-etal-2020-neural}.

These results have encouraged interest in deploying models in domains like medicine and child protective services, where a small number of practitioners need to quickly obtain information from large volumes of text \citep{uzuner2012evaluating,Saxena2020Human}.
However,  successes over curated data sets have not fully translated to text containing technical vocabulary, frequent typos, or inconsistent syntax.
Coreference models struggle to produce meaningful representations for new domain-specific spans and may require many examples to adapt \citep{uppunda-etal-2021-adapting, lu-ng-2020-conundrums,zhu-etal-2021-anatomy}.

Further, coreference models trained on standard benchmarks are not robust to differences in annotation schemes for new domains \citep{bamman-etal-2020-annotated}. For example, OntoNotes does not annotate \textit{singleton} mentions, those that do not corefer with any other mention. A system trained on OntoNotes would implicitly learn to detect only entities that appear more than once, even though singleton retrieval is often desired in other domains \citep{zeldes}. Also, practitioners may only be interested in retrieving a subset of domain-specific entities.

Continued training on target domain data is an effective approach \citep{xia-van-durme-2021-moving}, but it requires costly and time-consuming coreference chain annotations in the new domain \cite{sachan2015active}.
Annotating data in high-stakes domains like medicine and child protective services is particularly difficult, where privacy needs to be preserved, and domain experts have limited time. 

Our work demonstrates that annotating only mentions is more efficient than annotating full coreference chains for adapting coreference models to new domains with a limited annotation budget.  First, through timed experiments using the i2b2/VA medical notes corpus \citep{uzuner2012evaluating}, we show that most documents can be annotated for mention detection twice as fast as for coreference resolution (\Sref{sec:timed-exp}).
Then, we propose how to train a coreference model with mention annotations by introducing an auxiliary mention detection objective to boost mention precision (\Sref{sec:methods}).

With this auxiliary objective, we observe that fewer antecedent candidates yields stronger linker performance. Continuity with previous feature-based approaches \citep{moosavi-strube-2016-search, recasens-etal-2013-life, wu-gardner-2021-understanding} suggests this relationship between high-precision mention detection and strong coreference performance in low-resource settings extends beyond the  architecture we focus on \citep{lee-etal-2018-higher}.  
% Consistent with previous feature-based approaches \citep{moosavi-strube-2016-search, recasens-etal-2013-life, wu-gardner-2021-understanding}, we observe that fewer antecedent candidates yields stronger linker performance, suggesting that high precision mention detection is especially critical in low-resource settings. 

We evaluate our methods using English text data from three domains: OntoNotes \citep{pradhan-etal-2012-conll}, i2b2/VA medical notes \citep{uzuner2012evaluating}, a new (unreleased) corpus of child welfare notes obtained from a county-level Department of Human Services (DHS).
We experiment with standard benchmarks for reproducibility, but we focus primarily on real-world settings where there is interest in deploying NLP systems and limited capacity for in-domain annotations \citep{uzuner2012evaluating,Saxena2020Human}.
For a fixed amount of annotator time, our method consistently out-performs continued training with target domain coreference annotations when transferring both within or across annotation styles and vocabulary. 

% In summary, we show that standard continued training approaches to adapting coreference resolution models with coreference examples from the target domain are not the most efficient use of annotator time. In fact, adapting mention detection is often more effective in low-resource settings (\Sref{sec:small_experiment}). Through timing experiments, we show that mention detection is faster to annotate (\Sref{sec:timed-exp}) and easy to integrate (\Sref{sec:methods}) into a common coreference architecture \citep{lee-etal-2018-higher}.

Our primary contributions include: Timing experiments showing the efficiency of mention annotations (\Sref{sec:timed-exp}), and methodology to easily integrate mention annotations (\Sref{sec:methods}) into a common coreference architecture \citep{lee-etal-2018-higher}. Furthermore, to the best of our knowledge, this is the first work to examine coreference resolution in child protective settings. 
With empirical results demonstrating 7-14\% improvements in F1 across 3 domains, we find that our approach for adaptation using mention annotations alone is an efficient approach for practical, real-world datasets.
% By showing adapting mention detection is often more effective in low-resource settings, our work broadens the accessibility of coreference systems for technical domain-specific text.
% emohasize that we are demonstrating that our approach works in a real setting -- works and very practical, straughtforward. and we experiment in a real setting

% Our primary contributions include, timed experiments demonstrating the efficiency of mention annotations, methodology for incorporating mention annotations into neural coreference models, and empirical results demonstrating 7-14\% improvements in F1 across 3 domains. Furthermore, to the best of our knowledge, this is the first work to examine coreference resolution in child protective settings. Our work has the potential to increase the usability of coreference systems in domain-specific technical settings.

% we show that current standard approaches to 

% We experiment with a corpus of child welfare notes that we developed, medical notes \citet{uzuner20112010}, and genres in OntoNotes \citet{pradhan-xue-2009-ontonotes}.

\section{Background and Task Definition}\label{sec:background}
% In this section, we define our task setup: domain adaptation for coreference resolution. We then provide background information on neural coreference models and motivational evidence of our underlying hypothesis: that the mention detector is the most critical component to adapt to new domains. 

\subsection{Neural Coreference Models}
We focus our examination on the popular and successful neural approach to coreference introduced in \citet{lee-etal-2017-end}. This model includes three components: an encoder to produce span representations, a mention detector that outputs mention scores for candidate mentions, and a linker that outputs candidate antecedent scores for a given mention.
For a document of length $T$, there are $\frac{T(T-1)}{2}$ possible mentions (sets of contiguous words).

For the set of candidate mentions, the system assigns a pairwise score between each mention and each candidate antecedent. The set of candidate antecedents is all previous candidate mentions in the document and a dummy antecedent (representing the case where there is no antecedent). For a pair of spans $i, j$, the pairwise score is composed of mention scores $\textit{s}_m(i), \textit{s}_m(j)$ denoting the likelihood that spans $i$ and $j$ are mentions and an antecedent score $\textit{s}_a(i,j)$ representing the likelihood that span $j$ is the antecedent of span $i$.
\begin{align*}
    \textit{s}(i,j) = \textit{s}_m(i) + \textit{s}_m(j) + \textit{s}_a(i,j) 
\end{align*}

% describe different approaches to mention pruning
This architecture results in model complexity of $O(T^4)$, so it is necessary to prune the set of mentions.
\citet{lee-etal-2018-higher} introduce coarse-to-fine (c2f) pruning: of $T$ possible spans, c2f prunes the set down to
$M$ spans based on span mention scores $\textit{s}_m(i)$. Then for each span $i$, we consider antecedent $j$ based on the sum of their mention scores $\textit{s}_m(i), \textit{s}_m(j)$ and a coarse but efficient pairwise scoring function as defined in \citet{lee-etal-2018-higher}. 

\subsection{Domain Adaptation Task Setup}
In this work we investigate the following pragmatic domain adaptation setting: Given a text corpus annotated for coreference from source domain $S$, an un-annotated corpus from target domain $T$, and a limited annotation budget, our goal is to maximize coreference F1 performance in the target domain under the given annotation budget. We define this budget as the amount of annotation time. 

The most straightforward approach to this task is to annotate documents with full coreference chains in the target domain until the annotation budget is exhausted. Given an existing coreference model trained on the source domain, we can continue training on the annotated subset of the target domain. With a budget large enough to annotate at least 100 documents, this has been shown to work well for some domains \citep{xia-van-durme-2021-moving}.

\subsection{Effect of In-Domain Training on Mention Detection and Antecedent Linking}\label{sec:small_experiment}
Given that out-of-domain vocabulary is a common aspect of domain shift in coreference models \citep{uppunda-etal-2021-adapting, lu-ng-2020-conundrums}, we hypothesize that mention detection transfer plays an important role in overall coreference transfer across domains. 
To test this hypothesis, we conduct a preliminary experiment, examining how freezing the antecedent linker affects overall performance in the continued training domain-adaptation setting described above.  We train a c2f model with a SpanBERT encoder \citep{joshi-etal-2020-spanbert} on OntoNotes, a standard coreference benchmark, and evaluate performance over the i2b2/VA corpus, a domain-specific coreference data set consisting of medical notes (see \S\ref{sec:experimental_setup} for details). We additionally use the training set of i2b2/VA for continued in-domain training, and we isolate the impact of mention detection by training with and without freezing the antecedent linker.

Results are given in \autoref{table:oracle}. Continued training of just the encoder and mention detector results in a large improvement of 17 points over the source domain baseline, whereas unfreezing the antecedent linker does not further significantly improve performance. This result implies that mention detection can be disproportionately responsible for performance improvements from continued training.
If adapting only the encoder and mention detection portions of the model yields strong performance gains, this suggests that mention-only annotations, as opposed to full coreference annotations, may be sufficient for adapting coreference models to new domains. %We elaborate on the details of mention-only annotations in the following sections.
% Further, we can verify that mention detection is in fact the lion's share of adaptation. We measure the performance improvement of continued training of the antecedent linker in the presence of GOLD mentions. Given GOLD mentions, continued training of the antecedent linker resultsGiven the set of GOLD mentions, we compare performance of our baseline c2f model to continued training for the antecedent linker. 

\begin{table}[!ht]
\centering
\scalebox{0.75}{
\begin{tabular}{lccl} 
\toprule
%\hline
% \multicolumn{4}{|c|}{Coreference System Component Transfer Ablation}  \\\hline
Model & Recall  & Precision  & F1 
% \\\hline
\\\midrule
% \\
SpanBERT + c2f	&	31.94	&	50.75	&	39.10	\\
+ tune Enc, MD only	    &	60.40	&	56.21	&	56.42	\\
+ tune Enc, AL, MD	&	60.51	&	57.33	&	56.71	\\
% + tune Enc, MD only	    &	60.40\twoS	&	56.21\threeS	&	56.42\threeS	\\
% + tune Enc, AL, MD	&	60.51\threeS	&	57.33\threeS	&	56.71\twoS	\\
\bottomrule
\end{tabular}}
\caption{When conducting continued training of a c2f model on target domain i2b2/VA, tuning the antecedent linker (AL) does not result in a significant improvement over just tuning the mention detector (MD) and encoder (Enc). All differences between tuned models and SpanBERT + c2f were statistically significant ($p$ < .05)}
\label{table:oracle}
\end{table}

% \begin{table}[!ht]
% \centering
% \scalebox{0.75}{
% \begin{tabular}
% {|c | c | c | c |} 
% \hline
% \multicolumn{4}{|c|}{Coreference System Component Transfer Ablation}  \\\hline
% Metric & R  & P  & F1 
% \\\hline
% SpanBERT + c2f	&	31.94	&	50.75	&	39.1	\\\hline\hline
% + tune Enc,MD only	    &	60.40	&	56.21	&	56.42\threeS	\\\hline
% % + tune AL,MD only	&	40.87	&	40.05	&	40.46	\\\hline
% % CEAFm	&	61.36	&	65.63	&	63.43	\\\hline
% + tune Enc,AL,MD	&	60.51	&	57.33	&	56.71\twoS	\\\hline	\end{tabular}}
% \caption{When conducting continued training of a cf2 model on target domain i2b2/VA, tuning the antecedent linker (AL) does not result in a significant improvement over just tuning the mention detector (MD) and encoder (Enc).}
% \label{table:oracle}
% \end{table}

\section{Timed Annotation Experiments}\label{sec:timed-exp}
In \S\ref{sec:background} we established that adapting just the mention detection component of a coreference model to a new domain can be as effective as adapting both mention detection and antecedent linking. In this section we demonstrate that annotating mentions is approximately twice as fast as annotating full coreference chains. While coreference has been established as a time-consuming task to annotate for domain experts \citep{aralikatte-sogaard-2020-model, li-etal-2020-neural}, no prior work measures the relative speed of mention versus full coreference annotation. Our results suggest, assuming a fixed annotation budget, coreference models capable of adapting to a new domain using only mention annotations can leverage a corpus of approximately twice as many annotated documents compared to models that require full coreference annotations.

We recruited 7 in-house annotators with a background in NLP to annotate two tasks for the i2b2/VA dataset. For the first mention-only annotation task, annotators were asked to highlight spans corresponding to mentions defined in the i2b2/VA annotation guidelines. For the second full coreference task, annotators were asked to both highlight spans and additionally draw links between mention pairs if coreferent.
All annotators used INCEpTION \citep{inception} and underwent a 45 minute training session to learn and practice using the interface before beginning timed experiments.\footnote{Annotators were compensated \$15/hr and applied for and received permission to access the protected i2b2/VA data.}

In order to measure the effect of document length, we sampled short (\textasciitilde200 words), medium (\textasciitilde500), and long (\textasciitilde800) documents. Each annotator annotated four documents for coreference resolution and four documents for mention identification (one short, one medium, and two long, as most i2b2/VA documents are long). Each document was annotated by one annotator for coreference, and one for mention detection. This annotation configuration maximizes the number of documents annotated (as opposed to the number of annotators per document), which is necessary due to the high variance in style and technical jargon in the medical corpus. In total 28 documents were annotated.
% \renewcommand{\arraystretch}{2}

% \begin{table}[!ht]
% \centering
% \scalebox{0.75}{
% \begin{tabular}
% {|c | c | c | c|} 
% \hline
% \multicolumn{4}{|c|}{Average Task Annotation Time (s)}  \\\hline
% Document Partition  & Coreference  & Mention & Speed-up
% \\\hline
% short (\textasciitilde200 words) &	287.3	& 186.1 & 1.54		\\\hline
% medium (\textasciitilde500 words) & 582.5 & 408.8 & 1.42 \\\hline
% long (\textasciitilde800 words)& 1306.1 & 649.5 & 2.01\\\hline\hline
% all & 881.2 & 475.9 & 1.85\\\hline
% \end{tabular}}
% \caption{Timed experiments of mention annotation as compared to full coreference annotations. Mention annotation 2X faster over longer documents.}
% \label{table:timed-exp}
% \end{table}

\begin{table}[!ht]
\centering
\scalebox{0.75}{
\begin{tabular}
{c  c  c  c} 
\toprule
\multicolumn{4}{c}{Average Task Annotation Time (s)}  \\\midrule
Document Partition  & Coreference  & Mention & Speed-up
\\\cmidrule(lr){1-1} \cmidrule(lr){2-2} \cmidrule(lr){3-3} \cmidrule(lr){4-4}
short (\textasciitilde200 words) &	287.3	& 186.1 & 1.54		\\
medium (\textasciitilde500 words) & 582.5 & 408.8 & 1.42 \\
long (\textasciitilde800 words)& 1306.1 & 649.5 & 2.01\\
all & 881.2 & 475.9 & 1.85\\\bottomrule
\end{tabular}}
\caption{Timed experiments of mention annotation as compared to full coreference annotations. Mention annotation 2X faster over longer documents.}
\label{table:timed-exp}
\end{table}

\autoref{table:exp-summary} reports the average time taken to annotate each document. On average it takes 1.85X more time to annotate coreference than mention detection, and the disparity is more pronounced (2X) for longer documents.
{In \autoref{table:timed-exp-agreement} (\autoref{app:additional_results}) we additionally report inter-annotator agreement. Agreement is slightly higher for mention detection, albeit differences in agreement for the two tasks are not significant due to the small size of the experiment, agreement is higher for mention detection.}

{Although results may vary for different interfaces, we show empirically that mention annotation is faster than coreference annotation.}
%Although results may vary for different interfaces, we show empirically that mention annotation is faster than coreference annotation.

% In order to effectively allocate resources to annotation for low-resource domains, we 
% In our model, we impose an additional layer of pruning $q$
% introduce end to end model
% focus on mention detection part
% 
\section{Model}\label{sec:methods}

Given the evidence that a large benefit of continued training for domain adaptation is concentrated in the mention detector component of the coreference system (\Sref{sec:small_experiment}), and that mention annotations are much faster than coreference annotations (\Sref{sec:timed-exp}), in this section, we introduce methodology for training a neural coreference model with mention annotations. Our approach includes two core components focused on mention detection: modification to mention pruning (\Sref{sec:mention_pruning}) and auxiliary mention detection training (\Sref{sec:aux_mention_loss}). We also incorporate an auxiliary masking objective (\Sref{sec:aux_mask}) targeting the encoder.

\subsection{Baseline}

In our baseline model architecture \citep{lee-etal-2018-higher}, model components are trained using a coreference loss, where $Y(i)$ is the cluster containing span $i$ predicted by the system, and $\text{GOLD}(i)$ is the GOLD cluster containing span $i$: 
\begin{align*}
    \textbf{CL} = \log \prod_{i=1}^\textit{N} \sum_{\hat{y} \in \mathcal{Y}(i) \cap \text{GOLD}(i)} \textit{P}(\hat{y})
\end{align*}\label{eq:coref_obj} Of the set of $N$ candidate spans, for each span $i$ we want to maximize the likelihood that the correct antecedent set $ \mathcal{Y}(i) \cap \text{GOLD}(i)$ is linked with the current span. The distribution over all possible antecedents for a given span $i$ is defined using the scoring function $s$ described in \Sref{sec:background}:
\begin{align*}
    \textit{P}(y) = \frac{e^{\textit{s}(i,y)}}{\sum_{y'\in\textit{Y}} e^{\textit{s}(i,y')}} 
\end{align*}
% To some extent, mention identification is implicitly learned since only $\text{GOLD}$ spans receive positive weight in the objective function $\textbf{CL}$. However, there is no positive weight for identification of singletons, so the mention detector in \citet{lee-etal-2017-end} learns anaphoric mention detection instead of mention detection. To address this limitation in domains that include singleton mentions, we can complement with a mention detection auxiliary objective to incentivize identification of all mentions (\Sref{sec:aux_mention_loss}). 
% This is a harder task \citep{wu-gardner-2021-understanding}, so in settings where there are few annotated target examples, the mention detector may conflate the two tasks resulting in poorer performance.

\subsection{Mention Pruning Modification}
\label{sec:mention_pruning}

As described in \S\ref{sec:background}, c2f pruning reduces the space of possible spans; however, there is still high recall in the candidate mentions. For example, our SpanBERT c2f model trained and evaluated over OntoNotes achieves 95\% recall and 23\% precision for mention detection.
% , .377 F1.
In state-of-the-art coreference systems, high recall with c2f pruning works well and makes it possible for the antecedent linker to correctly identify antecedents. Aggressive pruning can drop gold mentions.

Here, we hypothesize that in domain adaptation settings with a fixed number of in-domain data points for continued training, high-recall in mention detection is not effective. More specifically, it is evident that the benefits of high recall mention tagging are only accessible to highly discerning antecedent linkers. \citet{wu-gardner-2021-understanding} show that antecedent linking is harder to learn than mention identification, so given a fixed number of in-domain examples for continued training, the performance improvement from mention detection would surpass that of the antecedent linker. In this case, it would be more helpful to the flailing antecedent linker if the mention detector were precise. 

Based on this hypothesis, we propose \textit{high-precision c2f pruning} to enable adaptation using mention annotations alone. We  impose a threshold $q$ on the mention score $\textit{s}_m(i)$ so that only the highest scoring mentions are preserved. 

\subsection{Auxiliary Mention Detection Task}
\label{sec:aux_mention_loss}

We further introduce an additional cross-entropy loss to train only the parameters of the mention detector, where $x_i$ denotes the span representation for the $i$'th span produced by the encoder:  
\begin{align*}
    \textbf{MD} = - \sum_{i=1}^\textit{N}&g(x_i)\log\left(s_m(x_i)\right) \\&+ \left(1-g(x_i)\right)\log\left(1-s_m(x_i)\right) 
\end{align*}
The loss is intended to maximize the likelihood of correctly identifying mentions where the indicator function $g(x_i)=1$ iff $x_i$ is a GOLD mention. The distribution over the set of mention candidates is defined using the mention score $s_m$. The mention detector is learned using a feed-forward neural network that takes the span representation produced by the encoder as input. The mention identification loss requires only mention labels to optimize.

\subsection{Auxiliary Masking Task}\label{sec:aux_mask}
We additionally use a masked language modeling objective (\textbf{MLM}) as described in \citet{devlin-etal-2019-bert}. We randomly sample 15\% of the WordPiece tokens to mask and predict the original token using cross-entropy loss. This auxiliary objective is intended to train the encoder to produce better span representations.
Since continued training with an $\textbf{MLM}$ objective is common for domain adaptation \citet{gururangan-etal-2020-dont}, we also include it to verify that optimizing the \textbf{MD} loss is not implicitly capturing the value of the \textbf{MLM} loss.

\section{Experiments}
We evaluate our model on transferring between data domains and annotation styles.
To facilitate reproducibility and for comparison with prior work, we conduct experiments on two existing public data sets. We additionally report results on a new (unreleased) data set, which reflects a direct practical application of our task setup and approach.
% cite Stanford neural machine translation systems for spoken language domain.  for continued training

\subsection{Datasets}
\label{sec:datasets}
% \subsubsection{Source Domain}
\textbf{OntoNotes (ON)} (English) is a large widely-used dataset \citep{pradhan-etal-2012-conll} with standard train-dev-test splits. Unlike the following datasets we use, the annotation style excludes singleton clusters. OntoNotes is partitioned into genres: newswire (nw), Sinorama magazine articles (mz), broadcast news (bn), broadcast conversations (bc), web data (wb), telephone calls (tc), the New Testament (pt). 
% talk about genres

% \textit{Medical Notes}

\noindent\textbf{i2b2/VA Shared-Task (i2b2)} Our first target corpus is a medical notes dataset, released as a part of the i2b2/VA Shared-Task and Workshop in 2011 \citep{uzuner2012evaluating}. 
Adapting coreference resolution systems to clinical text would allow for the use of electronic health records in clinical decision support or general clinical research for example \citep{wang2018clinical}.
% In addition to coreference annotations, they include tags for the spans for categories like treatment, problem, test, person. 
The dataset contains 251 train documents, 51 of which we have randomly selected for development and 173 test documents.  
{The average length of these documents is 962.6 tokens with average coreference chain containing 4.48 spans.} 
The annotation schema of the i2b2 data set differs from OntoNotes, in that annotators mark singletons and only mentions specific to the medical domain (\textsc{problem}, \textsc{test}, \textsc{treatment}, and \textsc{person}).
% \textbf{Child welfare notes}
% We also develop a second target corpus over child welfare case notes from t

\noindent\textbf{Child Welfare Case Notes (CN)} Our second target domain is a new data set of contact notes from a county-level Department of Human Services (DHS).\footnote{{Upon the request of the department, we do not report the name of the county in order to preserve anonymity.}}  These notes, written by caseworkers and service providers, log contact with families involved in child protective services. 
% There are associated privacy concerns with handling such sensitive data addressed in \autoref{sec:ethical_concerns}.
Because of the extremely sensitive nature of this data, this dataset has not been publicly released.
However, we report results in this setting, as it reflects a direct, real-word application of coreference resolution and this work.
Despite interest in using NLP to help practitioners manage information across thousands of notes \citep{Saxena2020Human}, notes also contain domain-specific terminology and acronyms, and no prior work has annotated coreference data in this setting. While experienced researchers or practitioners can annotate a small subset, collecting a large in-domain data set is not feasible, given the need to preserve families' privacy and for annotators to have domain expertise.

Out of an initial data set of 3.19 million contact notes, we annotated a sample of 200 notes using the same annotation scheme as i2b2, based on conversations with DHS employees about what information would be useful for them to obtain from notes. {We adapt the set of entity types defined in the i2b2 annotation scheme to the child protective setting by modifying the definitions (\autoref{app:additional_results}, \autoref{table:concept-desc}).}
{To estimate agreement}, 20 notes were annotated by both annotators, achieving a Krippendorf's referential alpha of 70.5 and Krippendorf's mention detection alpha of 61.5 {(\autoref{app:additional_results}, \autoref{table:cn-agreement})}. 

On average, documents are 320 words with 13.5 coreference chains with average length of 4.7. We also replicated the timed annotation experiments described in \Sref{sec:timed-exp} over a sample of 10 case notes, similarly finding that it takes 1.95X more time to annotate coreference than mention detection.
We created train/dev/test splits of 100/10/90 documents, allocating a small dev set following \citet{xia-van-durme-2021-moving}.
% More details about can be found in \Sref{sec:anno}.

We experiment with different source and target domain configurations to capture common challenges with adapting coreference systems (\autoref{table:exp-summary}). We also select these configurations to account for the influence of singletons on performance metrics.
% \autoref{table:exp-summary} provides an overview of configurations.

% \begin{table}[!ht]
% \centering
% \scalebox{0.7}{
% \begin{tabular}
% {|c | c || c | c|} 
% \hline
% \multicolumn{4}{|c|}{Experiments}  \\\hline
% Source $S$ & Target $T$ & OOV Rate & Anno. Style Match \\\hline
% i2b2 & CN & $32.3\%$ & \checkmark \\
% ON & i2b2 & $20.8\%$ & \\
% ON $\text{Genre}_i$ & ON $\text{Genre}_j$ & ($8.1\%, 47.9\%$)&\checkmark\\\hline
% \end{tabular}}
% \caption{Summary of source-target configurations in our experiments. We experiment with transfer between domains with common or differing annotation style, where annotation style can dictate whether or not there are singletons annotated or domain-specific mentions to annotate for example.}
% \label{table:exp-summary}
% \end{table}

\begin{table}[!ht]
\centering
\scalebox{0.7}{
\begin{tabular}
{c  c  c  c} 
\toprule
% \multicolumn{4}{c}{Experiments}  \\\midrule
Source $S$ & Target $T$ & OOV Rate & Anno. Style Match \\
\midrule
% \cmidrule(lr){1-1} \cmidrule(lr){2-2} \cmidrule(lr){3-3} \cmidrule(lr){4-4} 
i2b2 & CN & 32.3\% & \checkmark \\
ON & i2b2 & 20.8\% & \\
ON $\text{Genre}_i$ & ON $\text{Genre}_j$ & (8.1\%, 47.9\%)&\checkmark\\\bottomrule
\end{tabular}}
\caption{Summary of source-target configurations in our experiments. We experiment with transfer between domains with common or differing annotation style, where annotation style can dictate whether or not there are singletons annotated or domain-specific mentions to annotate for example.}
\label{table:exp-summary}
\end{table}
\subsection{Experimental Setup}\label{sec:experimental_setup}

\paragraph{Baseline: c2f ($\textbf{CL}_S, \textbf{CL}_T$)} {For our baseline, we assume access to coreference annotations in target domain.} We use pre-trained SpanBERT for our encoder. In each experiment, we train on the source domain with coreference annotations optimizing only the coreference loss $\textbf{CL}_S$. Then, we continue training with $\textbf{CL}_T$ on target domain examples. 

{We additionally experiment with an alternative baseline (high-prec. c2f $\textbf{CL}_S, \textbf{CL}_T, \textbf{MD}_T$) in which coreference annotations are reused to optimize our $\textbf{MD}$ over the target domain. This allows for full utilization the target domain annotations. }

% our model should be in bottom row
% bottom one is proposed models, other things are other baselines
% move the CL + MD +MLM should be at the bottom or right under the baseline
% have another seprator (model ablations )

\begin{table*}[t!]
\centering
\small
\scalebox{0.8}{
\begin{tabular}
% {l c c *{10}{d{1.3}}} 
{l c c *{10}{S[table-format=1.2]}}%  *{9}{c{1.3}}} 

\toprule
\multirow{2}{*}{Model (\citet{lee-etal-2018-higher} + SpanBERT)}  & \multicolumn{2}{c}{Target Anno.}  & \multicolumn{5}{c}{ON$\to$i2b2} & \multicolumn{5}{c}{i2b2$\to$CN} \\ 
\cmidrule(lr){2-3} \cmidrule(lr){4-8} \cmidrule(lr){9-13}
% \\\cline{5-19}
& \mc{$\textbf{CL}_T$} & \mc{$\textbf{MD}_T$} & \mc{LEA} & \mc{MUC} & \mc{$\text{B}^3$} & \mc{$\text{CEAF}_{\phi}$} & \mc{Avg.} & \mc{LEA} & \mc{MUC} & \mc{$\text{B}^3$} & \mc{$\text{CEAF}_{\phi}$} & \mc{Avg.}\\
\rowcolor{lc0} + c2f ($\textbf{CL}_S, \textbf{CL}_T$)	&	$0\%$	& $0\%$	&	0.47	&	0.61	&	0.33	&	0.21	&	0.41	&	0.46	&	0.68	&	0.41	&	0.15	&	0.43	\\

\rowcolor{lc1} + c2f ($\textbf{CL}_S, \textbf{CL}_T$)\textsuperscript{\dag} &   25\% & 0\% &  0.65 & 0.75 & 0.44 & 0.29 & 0.53 & 0.49 & 0.70 & 0.42 & 0.16 & 0.44\\
\rowcolor{lc1} + high-prec. c2f ($\textbf{CL}_S, \textbf{MD}_T$) + Silver & 0\% & 50\% & 0.49\twoS & 0.63\threeS & 0.74\threeS & 0.61\threeS & \textbf{0.63}\threeS & 0.42\twoS & 0.70\threeS & 0.47\threeS & 0.22\threeS & \textbf{0.45}\oneS \\

\rowcolor{lc2} + c2f ($\textbf{CL}_S, \textbf{CL}_T$)\textsuperscript{\dag} & $50\%$ &	$0\%$		&	0.70	&	0.79	&	0.46	&	0.32	&	0.57	&	0.47	&	0.69	&	0.42	&	0.16	&	0.43	\\

\rowcolor{lc2} {+ high-prec. c2f ($\textbf{CL}_S, \textbf{CL}_T, \textbf{MD}_T$)\textsuperscript{\dag}} & $50\%$ & $0\%$ & 0.69 & 0.79 & 0.45 & 0.29 & 0.56 & 0.52 & 0.72 & 0.47 & 0.21 & 0.48\\

\rowcolor{lc2} + c2f ($\textbf{CL}_S, \textbf{MD}_T$) & 0\% & 100\% & 0.42\oneS & 0.56\threeS & 0.43 & 0.32 & 0.43 & 0.54\threeS & 0.77 &  0.47\oneS & 0.21\threeS & 0.49\oneS\\
\rowcolor{lc2} + high-prec. c2f ($\textbf{CL}_S, \textbf{MD}_T$)	& $0\%$ &	$100\%$	&	0.50\oneS	&	0.63\threeS	&	0.74\twoS	&	0.65	&	0.63\oneS	&	0.50	&	0.77\threeS	&	0.52	&	0.35\threeS	&	0.53	\\
\rowcolor{lc2} + high-prec. c2f ($\textbf{CL}_S, \textbf{MD}_T, \textbf{MLM}_T$)	& $0\%$ &	$100\%$	&	0.50\threeS	&	0.63\threeS	&	0.77\twoS	&	0.68\threeS	&	\textbf{0.64}\threeS	&	0.57\oneS	&	0.76\oneS	&	0.58	&	0.38	&	\bfseries{0.57}\oneS	\\

\rowcolor{lc3} + c2f ($\textbf{CL}_S, \textbf{CL}_T$) & $100\%$ &	$0\%$		&	0.71	&	0.80	&	0.48	&	0.33	&	0.58	&	0.77	&	0.86	&	0.63	&	0.29	&	0.64	\\
\bottomrule
% \hline	
\end{tabular}}
\captionof{table}{We report F1 for different models with singletons included in system output, varying the type and amount of target domain annotations. Each shade of gray represents a fixed amount of annotator time (e.g. 50\% Coreference and 100\% Mention annotations takes an equivalent amount of time to produce). With a limited annotation budget, for both the ON$\to$i2b2 and i2b2$\to$CN experiments, mention annotations are a more efficient use of time, yielding performance gains over the baseline with equivalent annotator time (i.e. indicated with {\dag}). \oneS denotes statistical significance with \textit{p}-value < .05}
\label{table:coref-perf-ws}
\end{table*}

\paragraph{Proposed: high-prec. c2f ($\textbf{CL}_S, \textbf{MD}_T, \textbf{MLM}_T$)}
We use the same model architecture and pre-trained encoder as the baseline, but also incorporate the joint training objective $\textbf{CL + MD}$. We optimize $\textbf{CL}$ with coreference examples from the source domain ($\textbf{CL}_S$), and $\textbf{MD}$ with examples from the target domain ($\textbf{MD}_T$). 
% We also more aggressively prune the space of mentions with threshold $q=.5$ on the mention score $s_m$ as described in \Sref{sec:methods}. 
We report results only with $\textbf{MD}_T$ paired with high-prec. c2f pruning (i.e. threshold $q=.5$ imposed on the mention score $s_m$)  as described in \Sref{sec:methods}. Without the threshold, $\textbf{MD}_T$ has almost no effect on overall coreference performance, likely because the space of candidate antecedents for any given mention does not shrink.

Our model uses only mentions without target domain coreference links, while our baseline uses coreference annotations. Accordingly, we compare results for settings where there is (1) an equivalent number of annotated documents and (2) an equivalent amount of annotator time spent, estimated based on the timed annotation experiments in \Sref{sec:timed-exp}.
% We additionally report results with and without \textbf{MLM} described in \Sref{sec:aux_mask}.

For each transfer setting, we assume the source domain has coreference examples allowing us to optimize $\textbf{CL}_S$. In the target domain, however, we are interested in a few different settings: (1) 100\% of annotation budget is spent on coreference, (2) 100\% of annotation budget is spent on mentions, (3) the annotation budget is split between mention detection and coreference. In the first and third settings we can optimize any subset of $\{\textbf{CL}_T,\textbf{MD}_T, \textbf{MLM}_T\}$ over the target domain, whereas $\textbf{CL}_T$ cannot be optimized for the second. 

We train the model with several different samples of the data, where samples are selected using a random seed. We select the number of random seeds based on the subsample size (\autoref{app:recproducibility}).

% \begin{itemize}
%     \item \textbf{i2b2} to \textbf{CN}: Both domains share identical annotation style (include singletons) as defined in the i2b2/VA shared task, but there is little vocabulary overlap.
%     \item \textbf{OntoNotes} to \textbf{i2b2}: The annotation style in OntoNotes is different from i2b2. I2b2 contains singletons and selectively annotates only specific types of entities
%     \item Genre X to Genre Y (\textbf{OntoNotes}): The annotation style between every pair of genres in OntoNotes is exactly identical (exclude singletons)
% \end{itemize}
% \subsection{Model Specifications}

% % genre-to-genre experiments
\subsection{Augmented Silver Mentions}
% Mention detection can be learned with fewer examples (\todo{reference appendix figure with examples to learn mention detection over the datasets}).
To further reduce annotation burden, we augment the set of annotated mentions over the target domain. 
We train a mention detector over a subset of gold annotated target-domain. Then, we use it to tag silver mentions over the remaining unlabeled documents, and use these silver mention labels in computing $\textbf{MD}_T$. 

\subsection{Coreference Evaluation Configuration}
In addition to the most common coreference metrics $\text{MUC}, \text{B}^3, \text{CEAF}_{\phi_4}$, we average across link-based metric $\text{LEA}$ in our score. We also evaluate each model with and without singletons, since including singletons in the system output can artificially inflate coreference metrics \citep{kubler-zhekova-2011-singletons}. When evaluating with singletons, we keep singletons (if they exist) in both the system and $\text{GOLD}$ clusters. When evaluating without singletons, we drop singletons from both.

\section{Results and Analysis}

% \input{tables/main-coref}
% state that overall performance improves with mentions
\autoref{table:coref-perf-ws} reports results when transfering models trained on ON to i2b2 and models trained on i2b2 to CN with singletons included (for completeness \autoref{app:additional_results}, \autoref{table:coref-perf-wos} reports results without singletons). 
For both i2b2$\to$CN and ON$\to$i2b2, our model performs better with mention annotations than the continued training baseline with half the coreference annotations (e.g. equivalent annotator time, since the average length of i2b2 documents is 963 words; and timed experiments in CN suggested mention annotations are \textasciitilde$2$X faster than coreference, \Sref{sec:datasets}). Combining $\textbf{MLM}_T$ with $\textbf{MD}_T$ results in our best performing model, but introducing $\textbf{MD}_T$ with high-precision c2f pruning is enough to surpass the baseline.
The results suggest in-domain mention annotation are more efficient for adaptation than coreference annotations.

\begin{figure}[htb!]
  \centering
  \includegraphics[width=0.4\textwidth]{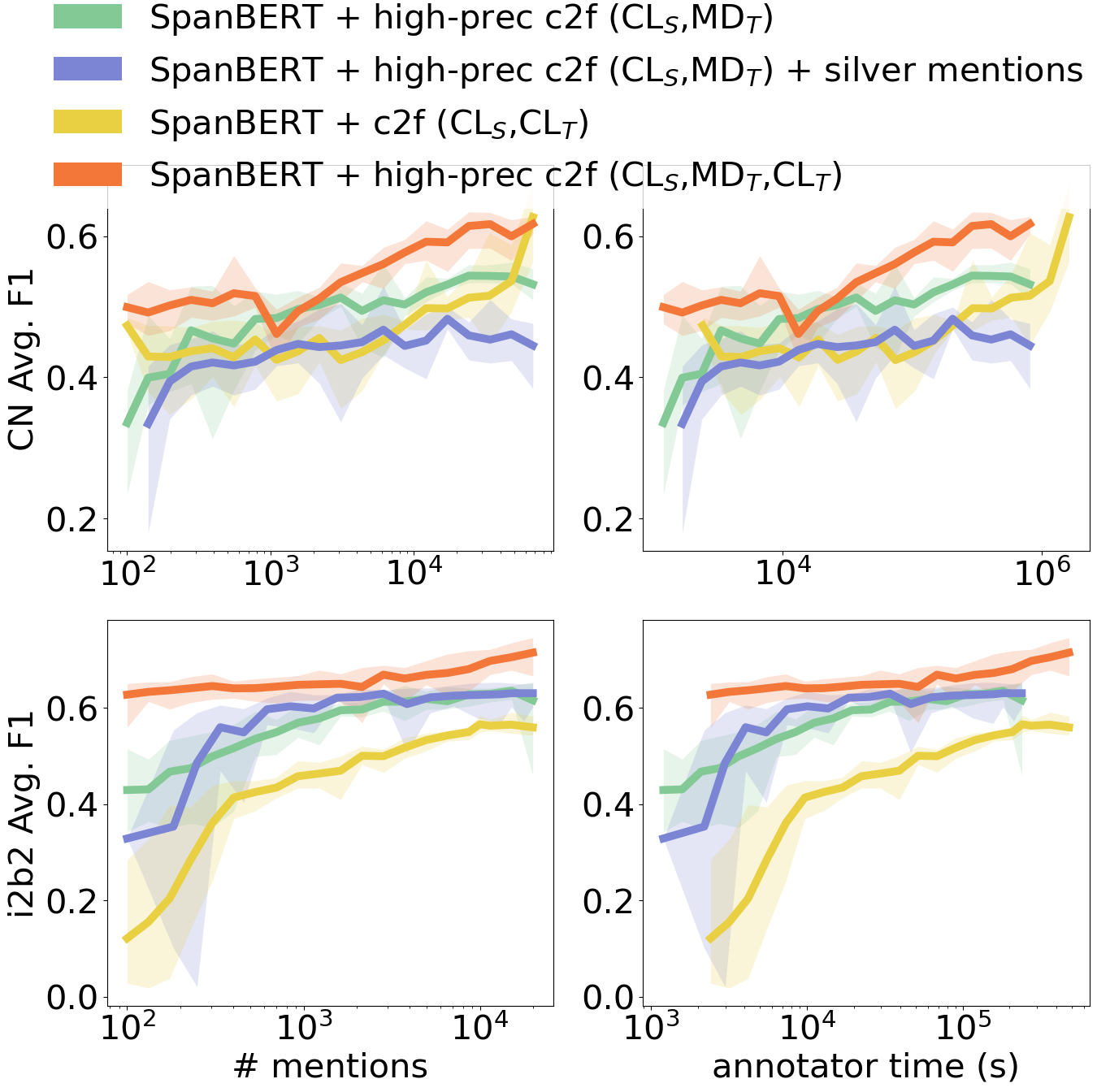}
    \caption{Each subplot shows coreference performance (singletons included) with varied amounts of annotated target domain data wrt the number of mentions (left) and the amount of annotator time (right). {Note that for ($\textbf{CL}_S,\textbf{MD}_T,\textbf{CL}_T$), we vary only the amount of coreference annotations -- the model accesses $100\%$ of mention annotations.} For ON$\to$i2b2 (bottom), our model ($\textbf{CL}_S, \textbf{MD}_T$) has the largest improvement over the baseline ($\textbf{CL}_S, \textbf{CL}_T$) with limited  annotations/time. For the i2b2$\to$CN (top), however, the disparity increases with more annotations.}
    \label{fig:i2b2-cn-exp}

  \vspace{1ex}

  \includegraphics[width=.45\textwidth]{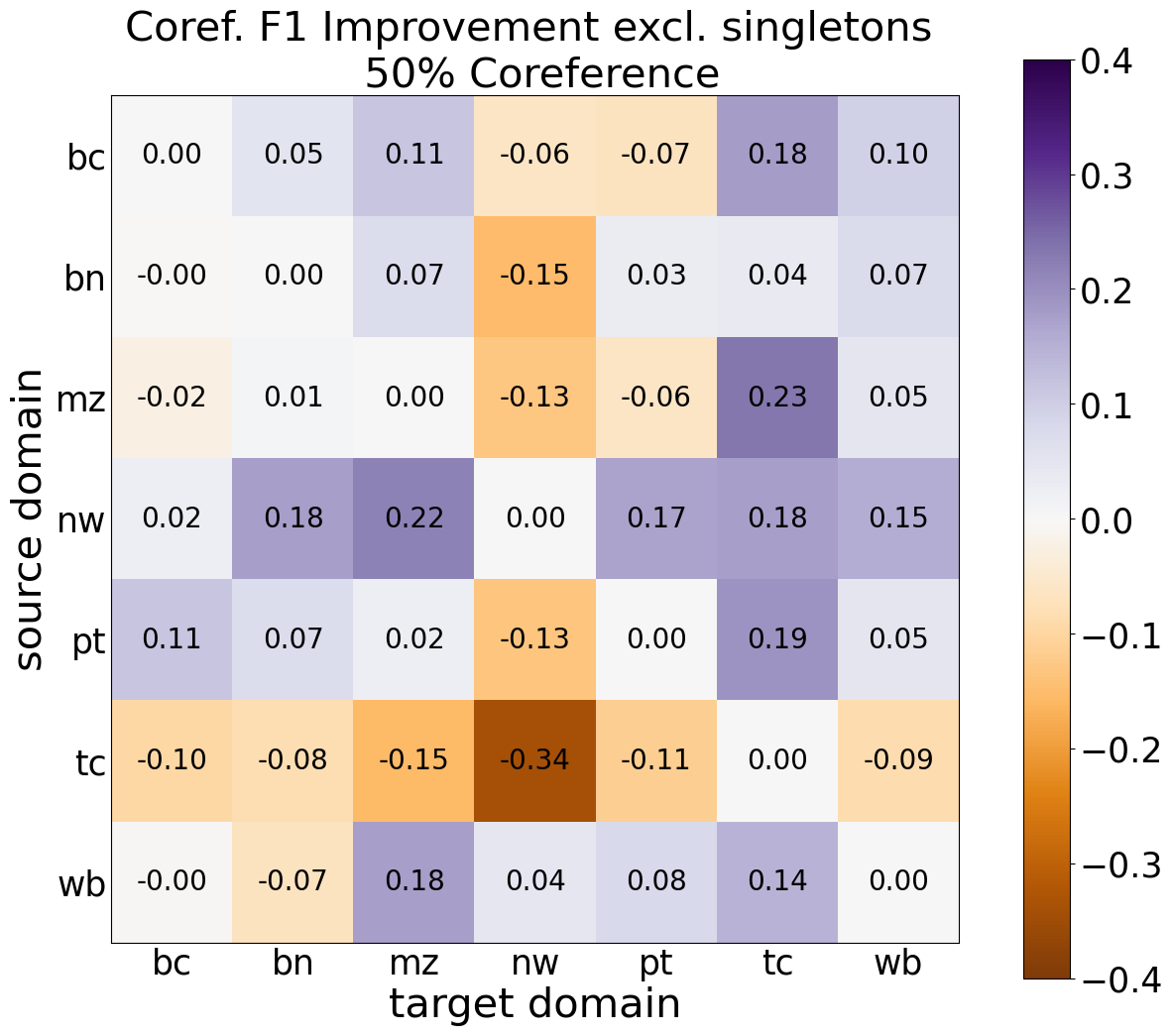}
    
    \caption{Heatmap represents performance improvements from our model where singletons are excluded. Our model SpanBERT + high-prec c2f ($\textbf{CL}_S,\textbf{MD}_T$) accesses 100\% mention annotations from the target domain, and the baseline SpanBERT + c2f ($\textbf{CL}_S,\textbf{CL}_T$) accesses 50\% of coreference examples. Annotating mentions for an equivalent amount of time is much more efficient for most ON genres. }
    \label{fig:g2g-combined}
\end{figure}
% \begin{figure}[ht]
% % \vspace{1ex}%
%     \centering
%     \includegraphics[width=0.4\textwidth]{images/i2b2_cn.png}
%     \caption{Each subplot shows coreference performance (singletons included) with varied amounts of annotated target domain data wrt the number of mentions (left) and the amount of annotator time (right). For ON$\to$i2b2 (bottom), our model ($\textbf{CL}_S, \textbf{MD}_T$) has the largest improvement over the baseline ($\textbf{CL}_S, \textbf{CL}_T$) with limited  annotations/time. For the i2b2$\to$CN (top), however, the disparity increases with more annotations.}
%     \label{fig:i2b2-cn-exp}
% \end{figure}
% \vspace{1ex}
% \begin{figure}[ht!]
%     \centering
%     \includegraphics[width=.45\textwidth]{images/combined-half-full-Average f1-50coref-diff.png}
    
%     \caption{Each subplot represents performance improvements from our model where singletons are excluded. Our model SpanBERT + high-prec c2f ($\textbf{CL}_S,\textbf{MD}_T$) accesses 100\% mention annotations from the target domain, and the baseline SpanBERT + c2f ($\textbf{CL}_S,\textbf{CL}_T$) accesses 50\% of coreference examples. Annotating mentions for an equivalent amount is much more efficient for most ON genres. }
%     \label{fig:g2g-combined}
% \end{figure}

\subsection{Transfer Across Annotation Styles}
ON and i2b2 have different annotation styles (\Sref{sec:experimental_setup}), allowing us to examine how effectively mention-only annotations facilitate transfer not just across domains, but also across annotation styles. Transferring ON$\to$i2b2 (\autoref{table:coref-perf-ws}), average F-1 improves by 6 points (0.57 to 0.63), when comparing the baseline model with 50\% coreference annotations with our model (i.e. equivalent annotator time). 

In \autoref{fig:i2b2-cn-exp} (top), we experiment with varying the amount of training data and annotator time in this setting. With more mentions, our model performance steadily improves, flattening out slightly after 1000 mentions. The baseline model continues to improve with more coreference examples. Where there is scarce training data (100-1000 mentions), mention annotations are more effective than coreference ones. This effect persists when we evaluate without singletons (\autoref{fig:i2b2-cn-exp-ns}). 

 The baseline likely  only identifies mentions that fit into the source domain style (e.g. \textsc{people}). Because the baseline model assigns no positive weight in the coreference loss for identifying singletons, in i2b2, entities that often appear as singletons are missed opportunities to improve the baseline mention detector. With enough examples and more entities appearing in the target domain as non-singleton, however, the penalty of these missed examples is smaller, causing the baseline model performance to approach that of our model. 
 % performance gap between the baseline model and our model to narrow.

\subsection{Silver Mentions Improve Performance}
From \autoref{fig:i2b2-cn-exp}, approximately 250 gold mentions are necessary for sufficient mention detection performance for silver mentions to be useful to our model. For fewer mentions, the mention detector is likely producing silver mention annotations that are too noisy. The benefit of access to additional data starts to dwindle around 3000 mentions.

\begin{figure}[ht!]
% \vspace{1ex}%
    \centering
    \includegraphics[width=0.45\textwidth]{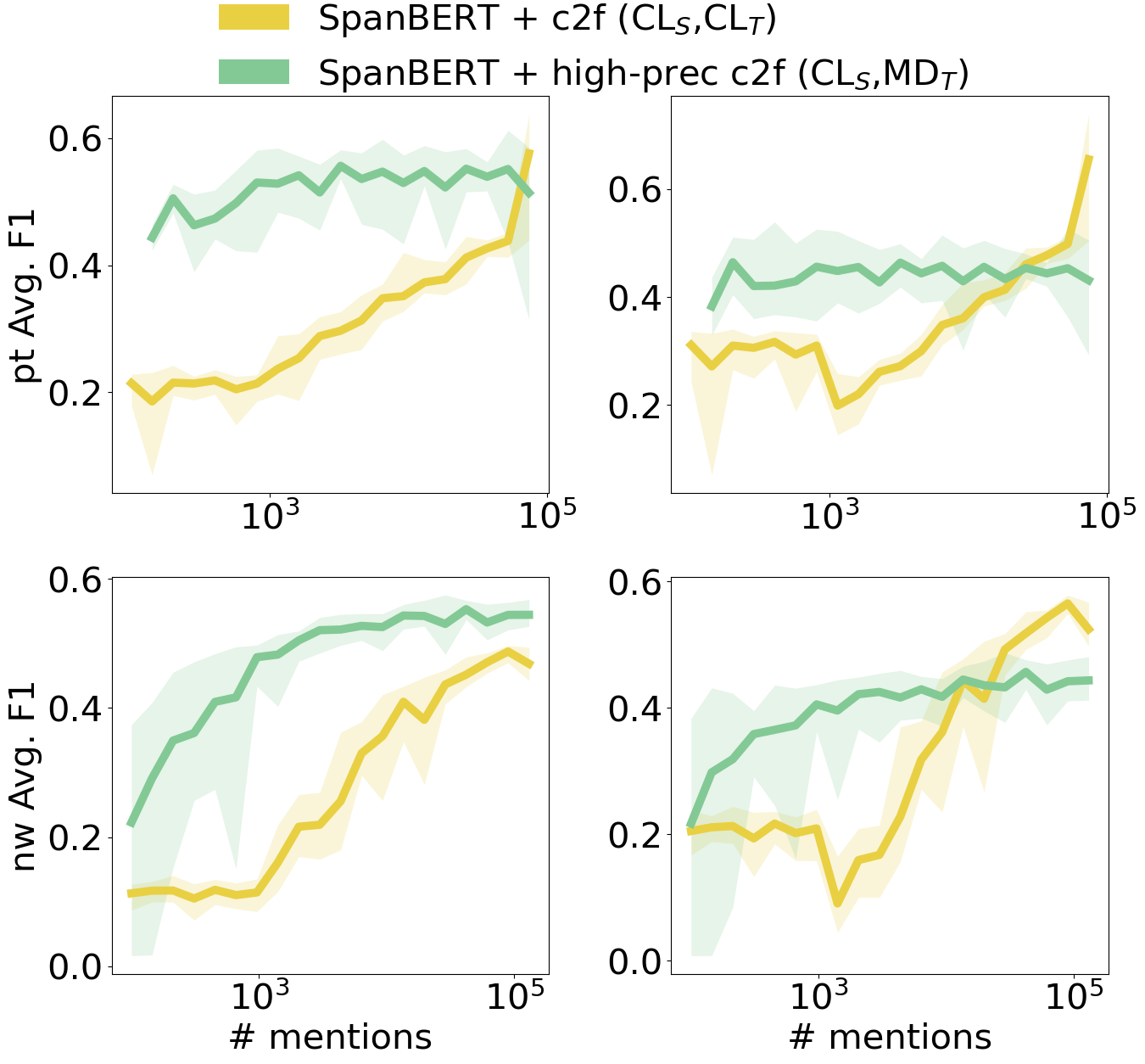}
    \caption{Each subplot shows coreference performance with varied amounts of annotated target data. 
    We report performance with singletons included in system output (left) and singletons excluded from system output (right) for two different genre-to-genre experiments: $bn\to pt$ (top) and $bn\to nw$ (bottom). 
    Regardless of whether singletons are included, annotating mentions is more efficient for all low-resource settings. 
    }
    \label{fig:bn-exp}
\end{figure}

\subsection{Fixed Annotation Style Transfer}
We additionally compare effects when transferring between domains, but keeping the annotation style the same.
When we transfer from i2b2 to CN, for equivalent annotator time, our model $\textbf{MD}_T + \textbf{MLM}_T$ improves over baseline $\textbf{CL}_T$ by 14 points (.43 to .57) in \autoref{table:coref-perf-ws}. (When singletons are dropped, this effect persists --- average F1 improves by 10 points, \autoref{app:additional_results}, \autoref{table:coref-perf-wos}). When we vary the number of mentions (\autoref{fig:i2b2-cn-exp}), the marginal benefit of CN mention annotations deteriorates for $>10^4$, but not as rapidly as when we transfer between annotation style in the ON$\to$i2b2 case. While mentions in CN share the same roles as those in i2b2, some types of mentions, (e.g. \textsc{problem}), are more difficult to identify.
Unlike settings where we transfer between annotation styles, when annotation style remains fixed, the performance improvement from our model increases with more target domain data. 
% \textcolor{red}{Further, given coreference annotations in the target domain, we find that reusing the annotations to optimize $\textbf{MD}$ with high-prec. c2f pruning can boost performance slightly when transferring within an annotation style. This is evident in the i2b2$\to$CN transfer setting regardless of whether singletons are included in the system output.} 
This suggests that adapting the mention detector is especially useful when transferring within an annotation style.

{Given coreference annotations, we find that reusing the annotations to optimize $\textbf{MD}_T$ with high-prec. c2f pruning boosts performance slightly when transferring within an annotation style.
This is evident in the i2b2$\to$CN case regardless of whether singletons are included in the output.}
% regardless of whether singletons are included in the system output. 

\autoref{fig:g2g-combined} reports results for the genre-to-genre experiments within ON. For equivalent annotator time our model achieves large performance improvements across most genres. Since our model results in significant improvements in low-resource settings when there are no singletons in the system or gold clusters, it is clear that performance gains are not dependent solely on singletons in the system output. 
% When the baseline $\textbf{CL}_T$ accesses 100\% coreference annotations (e.g. twice the annotator time), however, this improvement disappears for most source-target pairs (\autoref{fig:g2g-combined100}). 
% In the few cases where our model nevertheless performs better, it is possible that the antecedent linker more easily overfits than the mention detector component. For a given document, there tend to be more positive mention examples than positive linking examples.
\autoref{fig:bn-exp} shows varying the number of mentions and annotator time in settings where our model performed worse ($bn\to nw$) and better ($bn\to pt$) than the baseline. Regardless of transfer setting or whether singletons are excluded from the system output, our model out-performs the baseline with few mentions. 
\subsection{Impact of Singletons}
Under the with-singleton evaluation scheme, in the ON$\to$i2b2 case, the baseline trained with strictly more data performs worse than our model (\autoref{table:coref-perf-ws}, 0.58 vs. 0.64). \citet{kubler-zhekova-2011-singletons} describe how including singletons in system output causes artificial inflation of coreference metrics based on the observation that scores are higher with singletons included in the system output. Without high-precision c2f pruning with $\textbf{MD}_T$, the baseline drops singletons. So, the gap in \autoref{fig:i2b2-cn-exp} between the baseline  and our model at $10^4$ mentions could be attributed to artificial inflation. In the without-singleton evaluation scheme (\autoref{fig:bn-exp}, bottom) the artificial inflation gap between our model  and the baseline disappears with enough target examples, better reflecting our intuition that more data should yield better performance. But with fewer examples, our model still out-performs the baseline in the without-singleton evaluation scheme. 
% Further, independent of the presence of singletons, we can observe in \autoref{fig:g2g-combined} that annotating mentions is more efficient than coreference in low-resource settings.

In practical applications, such as identifying support for families involved in child protective services, retrieving singletons is often desired.
% For example, in the child welfare domain, there may be a grandmother mentioned once that could be a potential caregiver. Such infrequent mentions are more likely to be missed by a caseworker, and consequently more crucial to retrieve. 
Further, excluding singletons in the system output incentivizes high-recall mention detection, since the model is not penalized for a large space of candidate mentions in which valid mentions make up a small fraction. A larger space of possible antecedents requires more coreference examples to adapt antecedent linkers to new domains.

\section{Related Work}

Previous work has used  data-augmentation and rule-based approaches to adapt coreference models to new annotation schemes with some success \citep{toshniwal-etal-2021-generalization, zeldes-zhang-2016-annotation,https://doi.org/10.48550/arxiv.2205.12323}. 
In many cases, adapting to new annotation schemes is not enough -- performance degradation persists for out-of-domain data even under the same annotation scheme \citep{zhu-etal-2021-anatomy}, and encoders (SpanBERT)  can struggle to represent domain specific concepts well, resulting in poor mention recall \citep{timmapathini2021probing}. 

Investigation of the popular \citet{lee-etal-2017-end} architecture has found that coreference systems generally rely more on mentions than context \citep{lu-ng-2020-conundrums}, so they are especially susceptible to small perturbations. Relatedly, \citet{wu-gardner-2021-understanding} find that mention detection precision has a strong positive impact on overall coreference performance, which is consistent with findings on pre-neural systems \citep{moosavi-strube-2016-coreference, recasens-etal-2013-life} and motivates our work.

Despite challenges associated with limiting source domain annotation schema, with enough annotated data, coreference models can adapt to new domains. \citet{xia-van-durme-2021-moving} show that continued training is effective with at least 100 target documents annotated for coreference. However, it is unclear how costly it would be to annotate so many documents: while \citet{xia-van-durme-2021-moving} focus on the best way to use annotated coreference target examples, we focus on the most efficient way to spend an annotation budget.
% To address these challenges, domain knowledge can be effective in some cases. \citet{jindal2013using} uses domain knowledge extracted from UMLS and Wikipedia could be used to abstract the surface form of mentions for medical notes.  Further, curating the set of source examples \citet{aktas-etal-2020-adapting} or adding rules developed over the most challenging target examples \citet{uppunda-etal-2021-adapting} can be more effective than more data annotation.

A related line of work uses active learning to select target examples and promote efficient use of annotator time \citep{zhao-ng-2014-domain,li-etal-2020-active, yuan-etal-2022-adapting, miller-etal-2012-active}. 
% \citet{zhao-ng-2014-domain} introduced active learning as a mechanism for finding out-of-domain examples. There has been some work on optimizing annotation style in active learning for coreference. Samples are typically selected using an entropy level at a mention or cluster granularity \citep{li-etal-2020-active, yuan-etal-2022-adapting, miller-etal-2012-active}. 
% However, since these annotations require link information, there is a persistent trade-off in active learning between reading and labeling \citep{yuan-etal-2022-adapting}: If the sampling strategy confines samples to a single document, there will be redundant labeling effort. If the sampling strategy does not restrict the number of documents, there will be extra reading effort for the annotator. 
However, since these annotations require link information, there is a persistent trade-off in active learning between reading and labeling \citep{yuan-etal-2022-adapting}.
Since our method does not require link annotations for adaptation, our annotation strategy circumvents the choice between redundant labeling or reading. 
% Using our model, active learning approaches could collect more samples using the mention entropy function proposed in \citet{yuan-etal-2022-adapting}.

% There has been previous work focused on using alternative annotation strategies for coreference resolution. Observing that there is not a consensus defining the coreference task, \citet{dasigi-etal-2019-quoref} found that weaker notions of coreference can have benefits for downstream applications like question answering, machine translation \citep{poesio-etal-2019-crowdsourced}. 

% There has also been work to mitigate the mental burden of annotating coreference. \citet{aralikatte-sogaard-2020-model} designs the annotation as a mention-to-entity linking task instead of a mention-to-mention linking task to more closely match the mental model of a reader, speeding up the time to annotate. Our model, however, does not require coreference link annotations, greatly simplifying the annotation task.  

\section{Limitations}
Annotation speed for mention detection and coreference is dependent on many variables like annotation interface, domain expertise of annotators, annotation style, document length distribution. So, while our finding that coreference resolution is approximately 2X slower to annotate than mention detection held for two domains (i2b2, CN), there are many other variables that we do not experiment with.  
% mention/coref annotation speed is affected by many variables

We also experiment with transfer between domains with varying semantic similarity and annotation style similarity. But, our notion of annotation style is narrowly focused on types of mentions that are annotated (i.e. singletons, domain application-specific mentions). However, since our method is focused on mention detection, our findings may not hold for transfer to annotation styles with different notions of coreference linking (i.e. split-antecedent anaphoric reference \citep{yu-etal-2021-stay}). 
% notion of annotation style is narrow: mention types, singletons, but there are many other variations of coreference annotation style (e.g. split-antecedent anaphora), mention detection may not address that

We also focus on one common coreference architecture \citet{lee-etal-2018-higher} with encoder SpanBERT. However, there have been more recent architectures surpassing the performance of \citet{lee-etal-2018-higher} over benchmark ON \citep{dobrovolskii-2021-word, kirstain-etal-2021-coreference}. Our key finding that transferring the mention detector component can still be adopted.

\section{Ethical Concerns}\label{sec:ethical_concerns}
We develop a corpus of child welfare notes annotated for coreference. 
All research in this domain was conducted with IRB approval and in accordance with a data-sharing agreement with DHS.
Throughout this study, the data was stored on a secure disk-encrypted server and access was restricted to trained members of the research team.
Thus, all annotations of this data were conducted by two authors of this work.

While this work is in collaboration with the DHS, we do not view the developed coreference system as imminently deployable.
Prior to considering deploying, at a minimum a fairness audit on how our methods would reduce or exacerbate any inequity would be required. Deployment should also involve external oversight and engagement with stakeholders, including affected families.

\section{Conclusion}
Through timing experiments, new model training procedures, and detailed evaluation, we demonstrate that mention annotations are a more efficient use of annotator time than coreference annotations for adapting coreference models to new domains. Our work has the potential to expand the practical usability of coreference resolution systems and highlights the value of model architectures with components that can be optimized in isolation.

\section*{Acknowledgements}
Thanks to Yulia Tsvetkov, Alex Chouldechova, Amanda Coston, David Steier, and the %Allegheny County 
anonymous Department of Human Services for valuable feedback on this work. 
This work is supported by the Block Center for Technology and Innovation, and A.F. is supported by a Google PhD Fellowship.

\bibliography{anthology,custom}
\bibliographystyle{acl_natbib}

% gpus, 
% average runtime
% number of parameters
% evaluation on dev set

% focused on just one coref architecture, but key finding that mention detection precision helps in OOD settings holds -- but this only works when mention detector can be isolated
% estimate GPU time, we use RTX8000, v100, RTX600 on matcha (2900 hours)

% coref metrics may not reflect end-use application -- averaging coref metrics standard, but key limitations (e.g. artificial inflation from singletons)
% (revisit if they should be moved up later)
% (worth submitting the paper because there are reproducibility checklist -- go to submission page, there might be more stuff)
% doesn't super matter - will be desk rejected because of submission length

\appendix
% \newpage
\section{Additional Results}
\label{app:additional_results}

\begin{table*}[t!]
\centering
\small
\scalebox{0.8}{
\begin{tabular}
{l c c *{10}{S[table-format=1.2]}}
\toprule
\multirow{2}{*}{Model (\citet{lee-etal-2018-higher} + SpanBERT)}  & \multicolumn{2}{c}{Target Anno.}  & \multicolumn{5}{c}{ON$\to$i2b2} & \multicolumn{5}{c}{i2b2$\to$CN} \\ 
\cmidrule(lr){2-3} \cmidrule(lr){4-8} \cmidrule(lr){9-13}
% \\\cline{5-19}
& \mc{$\textbf{CL}_T$} & \mc{$\textbf{MD}_T$} & \mc{LEA} & \mc{MUC} & \mc{$\text{B}^3$} & \mc{$\text{CEAF}_{\phi}$} & \mc{Avg.} & \mc{LEA} & \mc{MUC} & \mc{$\text{B}^3$} & \mc{$\text{CEAF}_{\phi}$} & \mc{Avg.}\\
\rowcolor{lc0}+ c2f ($\textbf{CL}_S, \textbf{CL}_T$)	&	$0\%$	& $0\%$ &	0.47	&	0.61	&	0.49	&	0.24	&	0.45	&	0.46	&	0.68	&	0.49	&	0.38	&	0.50	\\
\rowcolor{lc1}+  c2f ($\textbf{CL}_S, \textbf{CL}_T$)\textsuperscript{\dag}  & 25\% & 0\% &  0.65 & 0.75\oneS & 0.68\oneS & 0.50 & \textbf{0.65}\oneS & 0.49 & 0.70 & 0.51 & 0.41 & \textbf{0.53}\\
\rowcolor{lc1}+ high-prec. c2f ($\textbf{CL}_S, \textbf{MD}_T$) + Silver & 0\% & 50\% & 0.49 & 0.63 & 0.50 & 0.15 & 0.44 & 0.42 & 0.70 & 0.44 & 0.23\oneS & 0.45 \\
\rowcolor{lc2}+ c2f ($\textbf{CL}_S, \textbf{CL}_T$)\textsuperscript{\dag} & $50\%$ &	$0\%$		&	0.70	&	0.79	&	0.72	&	0.57	&	\textbf{0.70}	&	0.47	&	0.69	&	0.50	&	0.40	&	0.51	\\

\rowcolor{lc2} {+ high-prec. c2f ($\textbf{CL}_S, \textbf{CL}_T, \textbf{MD}_T$)\textsuperscript{\dag}} & $50\%$ & $0\%$ & 0.69 & 0.79 & 0.72 & 0.57 & 0.69 &  0.52 & 0.72 & 0.55 & 0.45 & 0.56\\

\rowcolor{lc2} + c2f ($\textbf{CL}_S, \textbf{MD}_T$) & 0\% & 100\% & 0.42\oneS & 0.56 & 0.44 & 0.18 & 0.40\oneS & 0.54 & 0.77\oneS  & 0.56 & 0.45 & 0.58 \\

\rowcolor{lc2} + high-prec. c2f ($\textbf{CL}_S, \textbf{MD}_T$)	& $0\%$ &	$100\%$	&	0.50	&	0.63	&	0.53\oneS	&	0.32\oneS	&	0.49	&	0.50	&	0.77	&	0.52	&	0.42	&	0.55	\\
\rowcolor{lc2} + high-prec. c2f ($\textbf{CL}_S, \textbf{MD}_T, \textbf{MLM}_T$)	& $0\%$ &	$100\%$	&	0.50	&	0.63	&	0.51	&	0.22	&	0.47	&	0.57	&	0.76\oneS	&	0.60\oneS	&	0.49\oneS	&	\textbf{0.61}\oneS	\\	
\rowcolor{lc3}+ c2f ($\textbf{CL}_S, \textbf{CL}_T$) & $100\%$ &	$0\%$		&	0.71	&	0.80	&	0.74	&	0.61	&	0.71	&	0.77	&	0.86	&	0.78	&	0.71	&	0.78	\\
% SpanBERT + c2f ($\textbf{CL}_S, \textbf{CL}_T$)	&	$0\%$	& $0\%$	&	0.47	&	0.61	&	0.33	&	0.21	&	0.41	&	0.46	&	0.68	&	0.41	&	0.15	&	0.43	\\\cline{2-13}

\bottomrule

\end{tabular}}
\captionof{table}{We report F1 for different models with singletons excluded from system output, varying the type and amount of target domain annotations. Each shade of gray represents a fixed amount of annotator time (e.g. 50\% Coreference and 100\% Mention annotations takes an equivalent amount of time to produce). When transferring annotation styles (ON$\to$i2b2), coreference annotations are a more efficient use of time, while when transferring within an annotation style (i2b2$\to$CN), mention annotations are more efficient, consistent with results where singletons are included in the system output. Baselines are indicated with {\dag} and \oneS denotes statistical significance with \textit{p}-value < .05}
\label{table:coref-perf-wos}
\end{table*}

% i2b2
% baseline 0.62
% baseline_mm4951 0.64
% cl_md_ns_high_f1_wsilver_mm7094 0.37
% baseline_mm8646 0.68
% cl_md_ns_high_f102 0.17 
% cl_md_mlm_ns_high_f102 0.49 
% cl_md_ns_high_r 0.40
% baseline_mm19952 0.71
% cn
% cl
% baseline_i2b2_to_cn_mm17147
% cl_md_ns_high_f102_i2b2_to_cn_wsilver_mm34049__numseeds=6_ave
% baseline_i2b2_to_cn_mm34049
% cl_md_ns_high_f102_i2b2_to_cn
% cl_md_mlm_ns_high_f102
% cl_md_ns_high_r_i2b2_to_cn
% baseline_i2b2_to_cn

\begin{figure}[ht]
% \vspace{1ex}%
    \centering
    \includegraphics[width=0.4\textwidth,height=.533\textwidth]{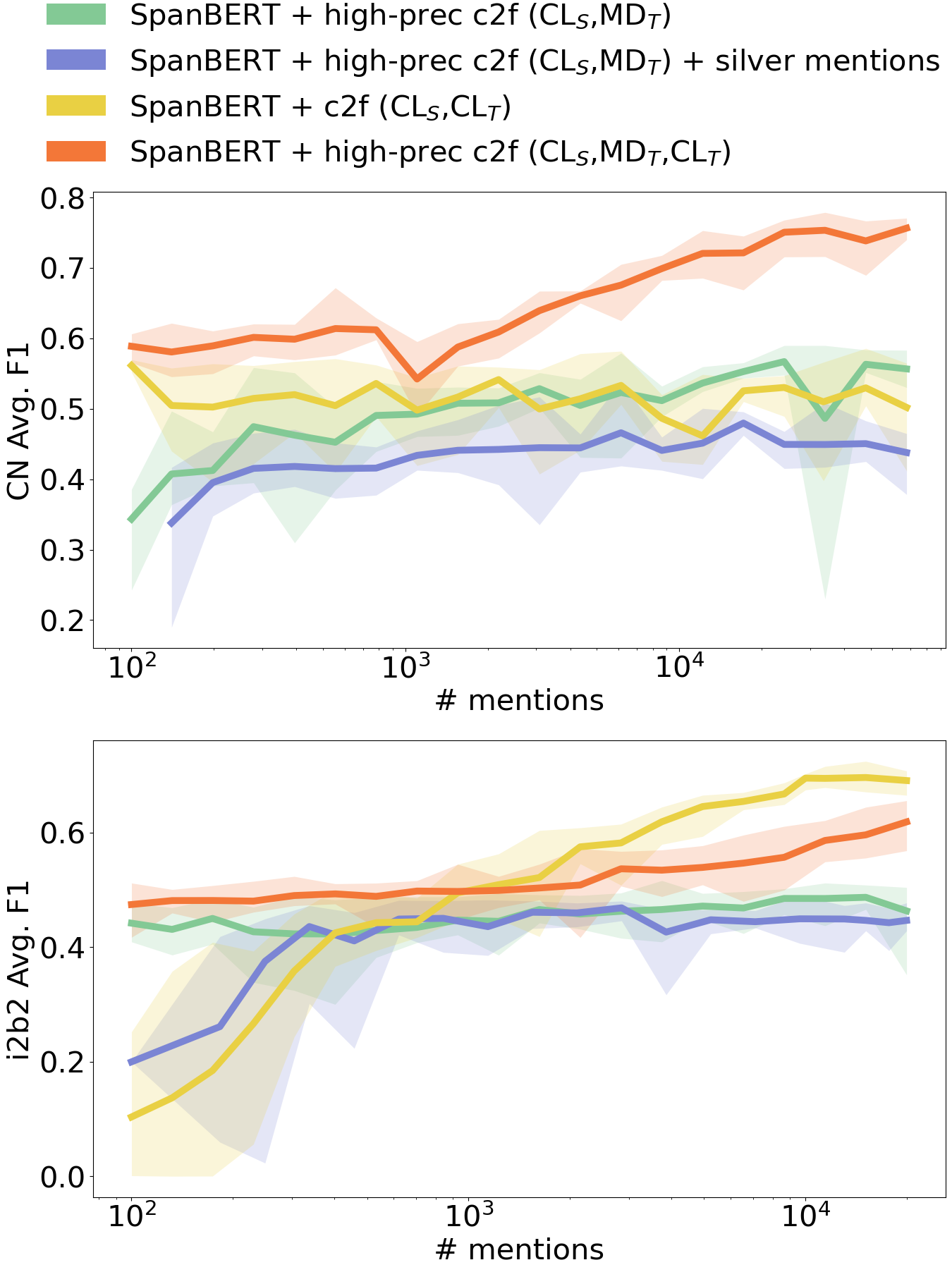}
    \caption{Each subplot shows coreference performance (singletons excluded) when trained with different amounts of annotated target domain data. We vary the amount of annotated data with respect to the number of mentions. When transferring ON$\to$i2b2 (bottom row), our model ($\textbf{CL}_S, \textbf{MD}_T$) has the largest improvement over the baseline ($\textbf{CL}_S, \textbf{CL}_T$) with very little training data or annotator time. For the i2b2$\to$CN (top row), however, the performance improvement increases with more annotated data.}
    \label{fig:i2b2-cn-exp-ns}
\end{figure}

\begin{figure}[ht!]
    \centering
    \includegraphics[width=.4\textwidth]{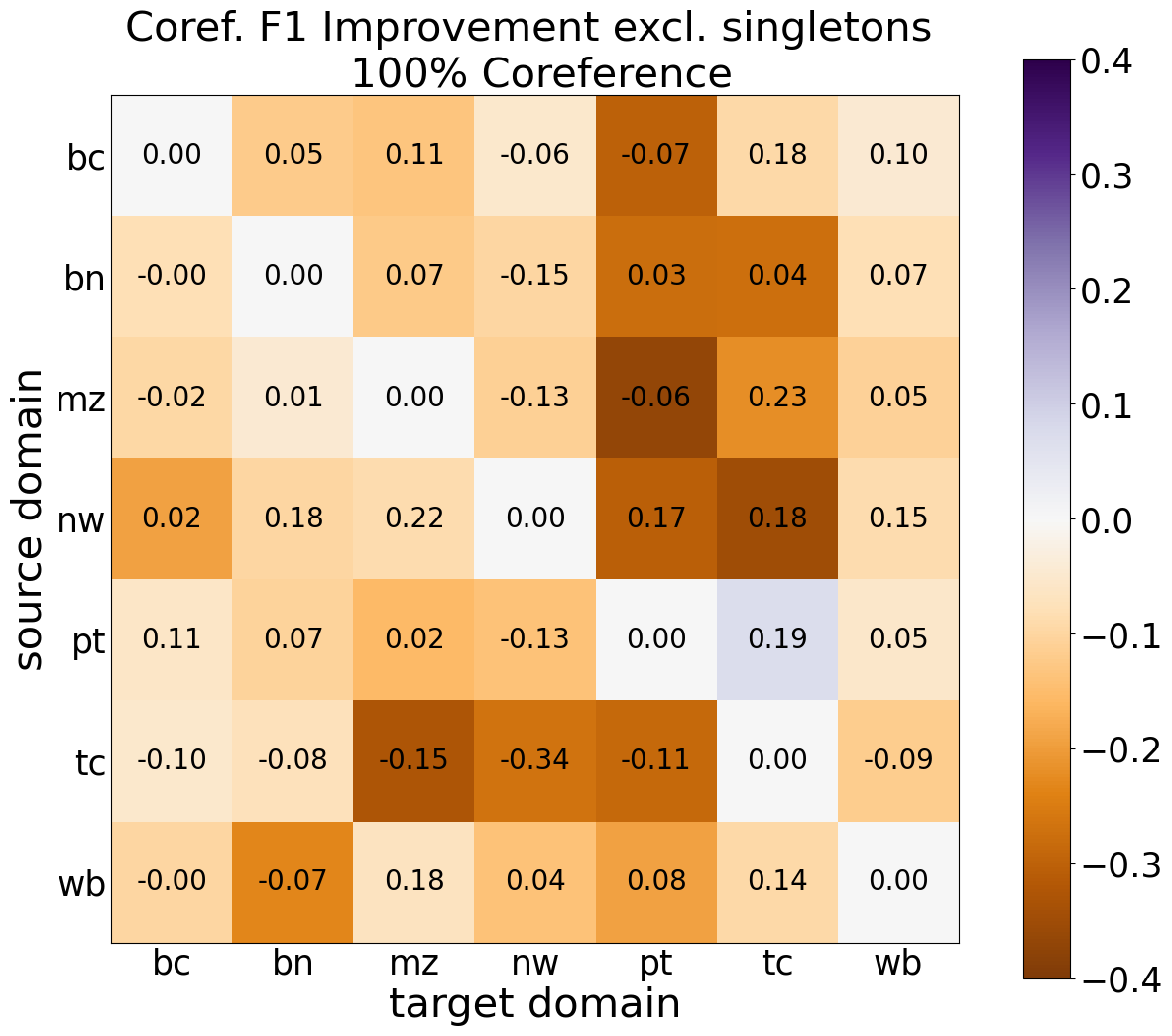}
    
    \caption{Heatmap represents performance improvements from our model SpanBERT + high-prec c2f ($\textbf{CL}_S,\textbf{MD}_T$) over the baseline SpanBERT + c2f ($\textbf{CL}_S,\textbf{CL}_T$) where singletons are dropped from the system output. The baseline has access to 100\% of target domain coreference examples, and our model has access to 100\% mention annotations.  }
    \label{fig:g2g-combined100}
\end{figure}

% now running baselinemd_i2b2_to_cn_mm34049_rs0
For completeness, we additionally include results with singletons omitted from system output. Table~\ref{table:coref-perf-wos} reports results for both transfer settings i2b2$\to$CN and ON$\to$i2b2. In Figure~\ref{fig:i2b2-cn-exp-ns}, we inspect how performance changes with more annotated data. 
We also report for completeness the difference in model performance using mention annotations and full coreference annotations in Figure~\ref{fig:g2g-combined100} for transfer between OntoNotes genres with an equivalent amount of annotated data (unequal amount of annotator time).

For our timed annotation experiment described in \Sref{sec:timed-exp}, we report more detailed annotator agreement metrics for the two annotation tasks in Table~\ref{table:timed-exp-agreement}. We expect that agreement scores for both tasks are low, since i2b2/VA dataset is highly technical, and annotators have no domain expertise. The increased task complexity of coreference resolution may further worsen agreement for the task relative to mention detection. We do not use this annotated data beyond timing annotation tasks.
% and Table~\ref{table:timed-exp-agreement-coref}.
% \input{tables/dev}

\begin{table}[!ht]
% \centering
\scalebox{0.75}{
\begin{tabular}
{l  c  c  } 
\toprule
\multicolumn{3}{c}{Timed Annotation Experiment Mention Detection Agreement}  \\\midrule
Agreement Metric  & Non-expert   & Domain-expert  \\
& Annotators & Annotators
\\\cmidrule(lr){1-1} \cmidrule(lr){2-2} \cmidrule(lr){3-3} 
Krippendorf's alpha &	0.405	& 	-	\\
Average Precision & 0.702 & - \\
Average Recall & 0.437 & - \\
Average F1 & 0.527 & - \\
IAA & 0.691 & 0.97 
\end{tabular}}
\newline
\vspace*{0.3cm}
\newline
\scalebox{0.75}{
\begin{tabular}
{l  c  c  } 
\toprule
\multicolumn{3}{c}{Timed Annotation Experiment Coreference Agreement}  \\\midrule
Agreement Metric  & Non-expert   & Domain-expert  \\
& Annotators & Annotators \\\cmidrule(lr){1-1} \cmidrule(lr){2-2} \cmidrule(lr){3-3} 
Krippendorf's alpha &	0.371	& 	-	\\
Average Precision & 0.275 & -\\
Average Recall & 0.511 & -\\
Average F1 & 0.342 & -\\
IAA & 0.368 & 0.73 
\end{tabular}}
\caption{Annotation agreement metrics for timed experiments of mention detection and coreference resolution. Inter-Annotator Agreement (IAA) refers to a metric defined in \cite{uzuner2012evaluating}. For coreference, precision, recall, and F1 are averaged over standard metrics defined in \Sref{app:recproducibility}.}
\label{table:timed-exp-agreement}
\end{table}

\section{Reproducibility Details}
\label{app:recproducibility}
\paragraph{Implementation Details}
For all models, we began first with a pretrained SpanBERT (base) encoder \citep{joshi-etal-2020-spanbert} and randomly initialized parameters for the remaining mention detector and antecedent linking. We use 512 for maximum segment length with batch size of one document similar to \citet{lee-etal-2018-higher}. We first train the model with a coreference objective over the source domain $\textbf{CL}_S$, and then we train over the target domain with some subset of our objectives $\textbf{CL}_T, \textbf{MD}_T, \textbf{MLM}_T$

We do not weight auxiliary objectives, taking the raw sum over losses as the overall loss. When we train one objective over both the source and target domain (i.e. $\textbf{CL}_S, \textbf{CL}_T$), we interleave examples from each domain. For the $\textbf{CL}$ objective, initial experiments indicated that, for fewer than 1k target domain mentions, our baseline model performed better if we interleaved target and source examples. So, we interleave target and source examples with fewer than 1k mentions from the target domain.

For experiments where the number of mentions from the target domain varied, we randomly sampled documents until the number of mentions met our cap (truncating the last document if necessary). For a given number of mentions $m$, we generated models for $\min(\max(6,15000/m),15)$ random seeds. These bounds were selected based on preliminary experiments assessing deviation.

We use a learning rate of $2 \times 10^{-5}$ for the encoder and $1\times 10^{-4}$ for all other parameters. We train on the source domain for 20 epochs and on the target domain for 20 epochs or until coreference performance over the dev set degrades for two consecutive iterations. 
Training time for all models ranges between 80-120 minutes, depending on size of dataset. We used V100, RTX8000, and RTX600 GPUS for training. To reproduce the results in this paper, we approximate at least 1,500 hours of GPU time. All our models contain  \textasciitilde134M parameters, with 110M  from SpanBERT (base).  

% 49*6*1 + 49*1*1 + 200*4*1 + 200*2*1  
\paragraph{Evaluation} We evaluate with coreference metrics: $\text{MUC}, \text{B}^3, \text{CEAF}_{\phi_4}, \text{LEA}$ for the ON$\to$i2b2 and i2b2$\to$CN transfer settings and only $\text{MUC}, \text{B}^3, \text{CEAF}_{\phi_4}$ for ON genre transfer experiments, since these three are standard for OntoNotes. We report results with singletons included and excluded from system output. Our evaluation script can be found at \texttt{src/coref/metrics.py}. 

\paragraph{CN Dataset Additional Details \label{sec:cn_data_extra}}
%Here we provide additional details about the Child Welfare Case Notes (CN) Dataset. 
Table~\ref{table:concept-desc} lists the specific definitions for labels used by annotators in the CN dataset, as compared to the descriptions in the i2b2/VA dataset after which they were modeled. Table~\ref{table:cn-agreement} reports measures for inter-annotator agreement for the CN dataset, compared to agreement reported for coreference annotations in OntoNotes.
\begin{table}[!h]
\centering
\scalebox{0.75}{
\begin{tabular}
{l  c  c  } 
\toprule
\multicolumn{3}{c}{CN Annotation Agreement}  \\\midrule
Agreement Metric  & Non-expert Annotators  & OntoNotes 
\\\cmidrule(lr){1-1} \cmidrule(lr){2-2} \cmidrule(lr){3-3} 
MUC &	72.0	& 	68.4	\\
$\text{CEAF}_{\phi}$ & 40.5 & 64.4\\
$\text{CEAF}_{m}$ & 63.4 & 48.0\\
$\text{B}^3$ & 57.8 & 75.0 \\
Krippendorf's MD alpha & 60.5 & 61.9 \\
Krippendorf's ref. alpha & 70.5 & $-$ \\
\end{tabular}}
\caption{Annotation agreement metrics for the CN dataset computed over a random sample of 20 documents. We achieve agreement on par with OntoNotes \citep{pradhan-etal-2012-conll}.}
\label{table:cn-agreement}
\end{table}
\begin{table*}[ht!]

\centering \small
    \begin{tabular}{m{2cm}  p{6cm}  p{6cm} }
          &  i2b2/VA definition  & CN definition \\\hline\\
          
         \textsc{Treatment} & phrases that describe procedures, interventions, and substances given to a patient in an effort to resolve a medical problem (e.g. Revascularization, nitroglycerin drip) & phrases that describe efforts made  to improve outcome for child (e.g. mobile therapy, apologized) \\
         \textsc{Test} & phrases that describe procedures, panels, and measures that are done to a patient or a body fluid or sample in order to discover, rule out, or find more information about a medical problem (e.g. exploratory laproratomy, the ekg, his blood pressure) & phrases that describe steps taken to discover, rule out, or find more information about a problem (e.g. inquired why, school attendance) \\
        \textsc{Problem} & phrases that contain observations made by patients or clinicians about the patient’s body or mind that are thought to be abnormal or caused by a disease (e.g. new ss chest pressure, rigidity, subdued) & phrases that contain observations made by CW or client  about any client’s body or mind that are thought to be abnormal or harmful (e.g. verbal altercation, recent breakdown, lack of connection, hungry)\\
    \end{tabular}
    \caption{In addition to the \textsc{PERSON} entity type which is the same in both domains, we develop a set of types for the child welfare domain that can be aligned with those from the medical domain i2b2/VA as defined in \cite{uzuner2012evaluating}. While the development of these types were intended to facilitate transfer from the medical domain, they are not necessarily comprehensive or sufficiently granular for the downstream tasks that coreference systems may be used for in child protective settings. }
    \label{table:concept-desc}
\end{table*}

% However, since many of these systems based on \citet{lee-etal-2017-end} were developed over OntoNotes, anaphoricity resolution has been shown to also factor into out-of-domain coreference performance. Both  \citet{wu-gardner-2021-understanding} and  \citet{lu-ng-2020-conundrums} showed that anaphoricity resolution is a difficult task that mention detectors struggle to do alone. The difficulty of anaphoricity resolution is likely why the converse problem: identification of singletons has been shown to improve coreference performance  \citet{recasens-etal-2013-life}\todo{move to appendix}. 

\end{document}